\documentclass{sig-alternate}
\pdfoutput=1
\usepackage[ampersand]{easylist}
\usepackage[numbers]{natbib}
\usepackage{color,soul,framed,comment,graphicx,subcaption,bm}

\usepackage[colorlinks,citecolor=blue, linkcolor=blue]{hyperref}
\usepackage{times}
\newcommand{\E}{\text{E}} 

\newcommand{\mypar}[1]{\vspace{0.1in}\noindent \textbf{#1 \,}}

\newfont{\mycrnotice}{ptmr8t at 7pt}
\newfont{\myconfname}{ptmri8t at 7pt}
\let\crnotice\mycrnotice\let\confname\myconfname

\begin{document}
\let\crnotice\mycrnotice\let\confname\myconfname\permission{Permission to make digital or hard copies of all or part of this
work for personal or classroom use is granted without fee provided that copies
are not made or distributed for profit or commercial advantage and that copies
bear this notice and the full citation on the first page. Copyrights for
components of this work owned by others than the author(s) must be honored.
Abstracting with credit is permitted. To copy otherwise, or republish, to post
on servers or to redistribute to lists, requires prior specific permission
and/or a fee. Request permissions from Permissions@acm.org.}
\conferenceinfo{RecSys'15,}{September 16 - 20, 2015, Vienna, Austria}
\CopyrightYear{2015}
\crdata{978-1-4503-3692-5/15/09}

\title{Dynamic Poisson Factorization}
\subtitle{}

\numberofauthors{4} \author{
\alignauthor
Laurent Charlin\\
  \affaddr{Columbia University}\\
  \affaddr{New York, NY}\\
  \affaddr{lcharlin@cs.columbia.edu}
\alignauthor
Rajesh Ranganath\\
  \affaddr{Princeton University}\\
  \affaddr{Princeton, NJ}\\
  \affaddr{rajeshr@cs.princeton.edu}
\alignauthor
James McInerney\\
  \affaddr{Columbia University}\\
  \affaddr{New York, NY}\\
  \affaddr{james@cs.columbia.edu}
\and
\alignauthor
David M. Blei\\
  \affaddr{Columbia University}\\
  \affaddr{New York, NY}\\
  \affaddr{david.blei@columbia.edu}
}

\maketitle
\begin{abstract}
Models for recommender systems use latent factors to explain the
preferences and behaviors of users with respect to a set of items (e.g.,
movies, books, academic papers). Typically, the latent factors are assumed
to be static and, given these factors, the observed preferences and
behaviors of users are assumed to be generated without order. These assumptions limit
the explorative and predictive capabilities of such models, since users'
interests and item popularity may evolve over time. To address this, we propose dPF, a dynamic
matrix factorization model based on the recent Poisson factorization model
for recommendations. dPF models the time evolving latent factors with a
Kalman filter and the actions with Poisson distributions. We
derive a scalable variational inference algorithm to infer the latent factors.
Finally, we demonstrate dPF on 10 years of user click data from arXiv.org,
one of the largest repository of scientific papers and a formidable source
of information about the behavior of scientists. Empirically we show
performance improvement over both static and, more recently proposed,
dynamic recommendation models. We also provide a thorough exploration of
the inferred posteriors over the latent variables.
\end{abstract}

\makeatletter{}\section{Introduction}

\noindent

The modern internet provides unprecedented access to products and
information---examples include books, clothes, movies, news articles,
social media streams, and academic papers---but these choices
increasingly overwhelm its users. Recommender systems can alleviate
this problem.  Using historical behavior about what a user clicks (or
purchases or watches), recommender systems can learn the users'
preferences and form personalized suggestions for each.

Historical data accumulates. Today, services such as movie streaming
websites and academic paper repositories have had the same customers
for years, sometimes decades. During this time, user preferences
evolve. For example, a long-time fan of sports biographies might read
a science fiction novel that she finds inspiring, and then starts to
read more science fiction in place of biographies.  Similarly, items
may serve different audiences at different times.  For example,
academic papers become popular in different fields as one community
discovers the techniques of another; a 1950s paper from signal
processing might see renewed importance in the context of modern
machine learning.

The problem is that most widely-used recommendation methods assume
that user preferences and item attributes are
static~\citep{yi_beyond_2014,koren_ordrec_2011,ekstrand_collaborative_2011}.
Such methods would continue to recommend sports biographies to the new
fan of science fiction, and they might not notice that the 1950s paper
is now relevant to a new kind of audience.  This is the problem that
we address in this paper. We develop a new recommendation method,
dynamic Poisson factorization (dPF), that captures how users and items
change over time.

DPF is a factorization approach.  It represents users in terms of
latent preferences; it represents items in terms of latent attributes;
and it allows both preferences and attributes to change smoothly
through time.  As an example, we ran our algorithm on ten years of
data from arXiv.org, containing users clicking on computer science
articles.  (ArXiv.org is a preprint server of scientific articles that
has been serving users since 1991.  Our data span 2003--2013.)

\begin{figure*}[ht]
  \begin{center}
  \includegraphics[width=0.9\linewidth]{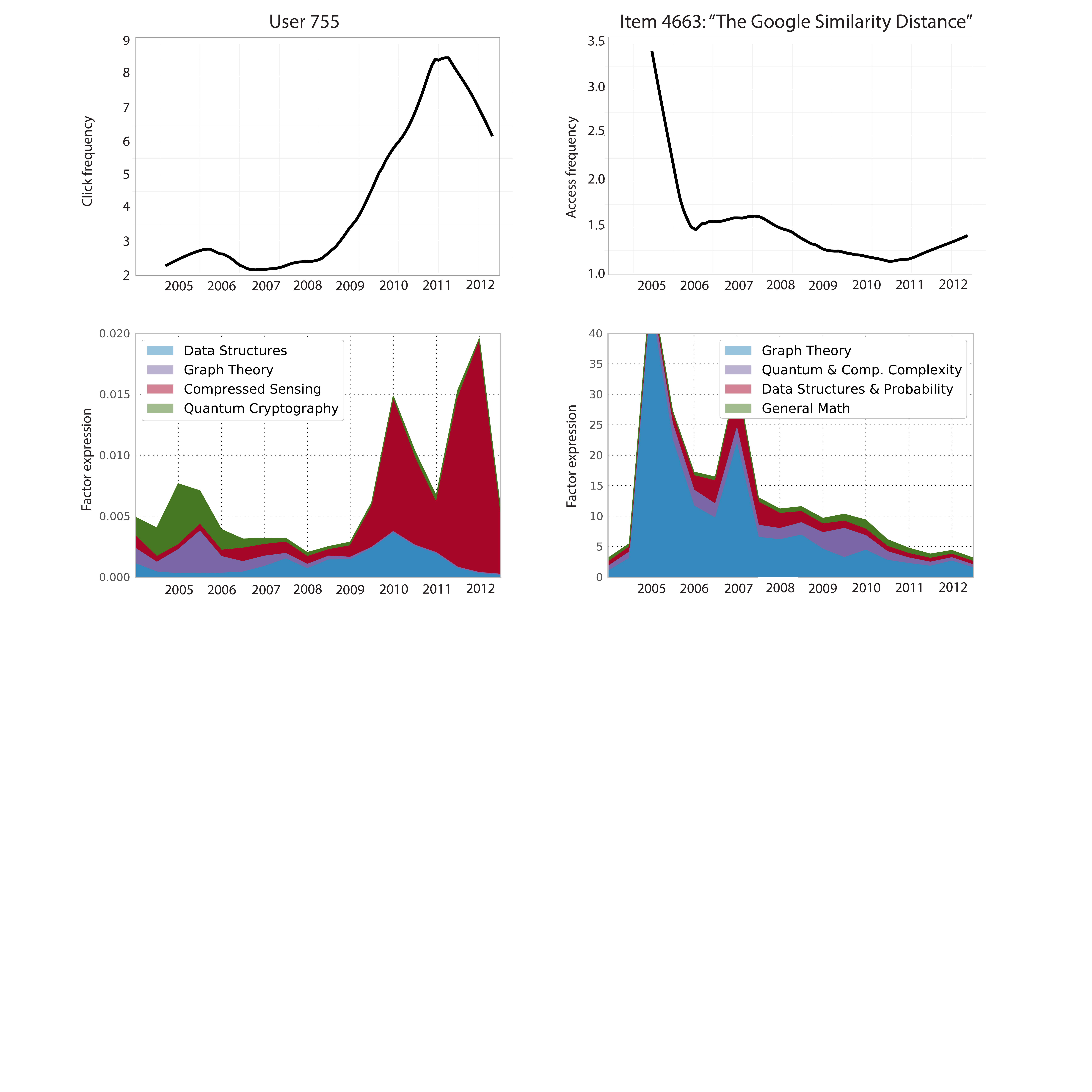}\vspace{-210pt}
          \end{center}
  \caption{Our method, dPF, discovers evolving user interests and item audiences over time from raw click data. 
The top left plot shows the aggregate click frequencies for a user on arXiv.org for 8 years. 
dPF separates this aggregate data into a set of 20 interest groups with varying strengths over time (bottom left). 
This user was initially interested in quantum cryptography and graph theory; then, five years later, in compressed sensing and data structures. DPF decomposes raw access counts for items in a similar way. The top right plot shows the raw access frequencies for the paper ``The Google Similarity Distance'' ({\small\url{http://arxiv.org/pdf/cs/0412098.pdf}}). The bottom right plot indicates that the paper broadened its audience during its lifetime. It was initially popular only with graph theorists, then received attention from quantum physicists and computational complexity readers, perhaps explained by the fact that it was cited in ``Google in a Quantum Network'' (on arXiv and then in the \emph{Nature} journal).}
  \label{fig:user755}
\end{figure*}

Figure~\ref{fig:user755} illustrates the kinds
of representations that it finds.
It shows the changing interests of an academic
reader. Ten years ago, this user was interested in graph theory and
quantum cryptography later the user was interested in data
structures; now the user is interested in compressed sensing  (For
convenience we have named the latent factors with meaningful labels.)
Note that a static recommendation engine would recommend graph theory
papers as strongly as compressed sensing papers, even though this user
lost interest in graph theory almost ten years ago. The figure also shows the latent attributes of the arXiv
paper ``The Google Similarity Distance'', and how those attributes
changed over the years.  At first it was popular with graph theorists;
since then it has found new audiences in math and quantum computing.
Again, a static recommendation system would not capture this change in
its audience.  Further, for both the user and the paper, the click
data alone (pictured on top in the figures) does not reveal the hidden
structure at play.

Dynamic Poisson factorization efficiently uncovers sequences of user
preferences and sequences of item attributes from large collections of
sequential click data, i.e., implicit data.  We used our algorithm to
study two large real-world data sets, one from Netflix and one from
arXiv.org. As we show in Section~\ref{sec:experiments}, the richer
representation of dPF leads to significantly improved recommendations.

\mypar{Technical approach. \,} Formally, dynamic Poisson factorization
is a new probabilistic model that builds on Poisson factorization
(PF)~\citep{Gopalan:2013b}.  PF is an effective model of implicit
recommendation data, and leads to scalable algorithms for computation
about large data sets.  Dynamic PF extends PF to handle time series of
clicks.  It mimics PF within each epoch, but allows the
representations of users and items to change across epochs.

While conceptually simple, this comes at the cost of some of the nice
mathematical properties of PF, specifically ``conditional conjugacy.''
Conditional conjugacy enables easy-to-derive algorithms for analyzing
data, i.e., for performing approximate posterior inference with
variational methods. Even without this property, we will derive an
efficient algorithm for dPF (Section~\ref{sec:model} and
Appendix~\ref{sec:inference}).  As with PF, our algorithm scales with
the number of non-zero entries in the user behavior matrix, enabling
large-scale analyses of sequential behavior data.

\makeatletter{}

\makeatletter{}
\begin{comment}
\begin{tabular}{p{1.25in}p{1.5in}}
\hline 
HMM & implicit + explicit data. \\ \hline 
Dynamic Bayesian Probabilistic Matrix Factorization & Gaussian observations, non-parametric user model over time \\ \hline 
Dynamic matrix factorization & Gaussian observations, user model over time, Kalman filtering for MAP inference \\ \hline 
Cross-Domain Collaborative Filtering over Time & extension of Bi-LDA over
time, matrix factorization with dirichlet-multinomials gen.
model. user \& item time model. Inference via sampling.  \\ \hline 
A collaborative Kalman filter for time-evolving dyadic processes & Gaussian observations. user and item time model. mean-field inference. \\ \hline 
Temporal Collaborative Filtering with Bayesian Probabilistic Tensor
Factorization & Data is encoded as a tensor. Time latent factors follow a
kalman-filter like evolution. \\ \hline 
TimesSVD++ (Koren) & MAP in a Gaussian model. Has global and
time-dependant factors but not as biases (not vectors) \\ \hline 
Nonparametric Bayesian Factor Analysis for Dynamic Count Matrices & Poisson observations. Non-parametric. Sampling-based inference \\ \hline 
\end{tabular}
\end{comment}

\mypar{Related Work.} Several authors emphasize the importance of
time modeling in collaborative filtering. These approaches differ in
how they incorporate time information and their use of implicit data.

\citet{DBLP:journals/cacm/Koren10} presents an approach based
on matrix factorization that models the ratings as a product of
user and item factors. This factorization is augmented with a user-item-time
bias and user evolving factors. Our work is similar, but we allow for item-evolving
factors to capture the possibility that item traits change over time (see Figure~\ref{fig:user755} for an example).
Further, we provide a probabilistic framework for our model with approximate
Bayesian inference to handle the considerable uncertainty that arises when
modeling time evolving user behaviors.  Similarly, Chatzis posits that only user membership
to a discrete \emph{group} (or class of users) evolves at each time-step, with new groups being 
created over time \cite{AAAI148136}.

Some methods propose factorization of the user-item-time
tensor~\citep{Karatzoglou:2010, lxiong:10:bptf}. The intuition behind these methods is to 
project user and item factors into click space using a time-specific factor. 
These methods, rather than modeling the evolution of users and items over 
time, quantify the activity of each dimension of the factor model over time. 
Thus comparing the trajectories of specific users and items is impossible.

\citet{GultekinP14} propose modeling both user and item evolution with a
Gaussian state-space model. For computational reasons, they perform a single forward (filtering) pass in time. 
In our work we show that we can perform both filtering and smoothing on large
datasets.  Similarly, \citet{Li:2011:CCF:2283696.2283780} use Dirichlet 
distributed latent factors, though they take a sampling-based inference approach that is not practical for real world datasets.

All the aforementioned methods are for explicit data.
Two additional methods focus on implicit data, though with only user evolving preferences. \citet{Sahoo:2012:HMM:2481674.2481689}
proposes an extension to the hidden Markov model where clicks are drawn from a negative binomial distribution. This discrete class-based 
approach requires many more classes to represent the
same size of space that our additive factors achieve. \citet{acharya15},
constructed from gamma processes, has the advantage of inferring
the dimensionality of the latent space, but takes an approach to inference
that is not scalable and does not allow for item evolution. 
Our work is the first model that considers implicit data 
where both the users and items can evolve over time. 

It is worth noting that recommendation frameworks to model general attributes,
including context such as location and time, have also been proposed
\cite{rendle:icdm10,Hidasi15}. In principle our work could be used to provide
predictions within the framweork of \citet{Hidasi15}.

\begin{easylist}

\end{easylist}

\makeatletter{}\section{Dynamic Poisson Factorization}\label{sec:model}

\noindent
In this section we review matrix factorization methods, Poisson matrix factorization,
and introduce dynamic Poisson factorization.

\mypar{Background: Matrix Factorization and Poisson Matrix Factorization.} 
\noindent The work of ~\citet{DBLP:journals/cacm/Koren10} is based on matrix 
factorization as are several of the other time-based methods highlighted at the start of this section. 
Matrix factorization aims at minimizing the distance between an
approximation of the matrix, $U^TV$, and the original matrix $Y$
($\text{dist}(Y,U^TV)$). This distance often has a
probabilistic interpretation. Therefore, obtaining a good approximation
(a small distance) can be cast as a maximum likelihood problem. For
example, obtaining an approximation to the user-item matrix in terms of
$l2$-loss is equivalent to assuming Gaussian noise on the observations.
Similarly, $l2$ regularization of the optimization variables is equivalent
to assuming Gaussian noise on the equivalent variables of a probabilistic
model \citep{bishop:2006:PRML}.

Expressing a model in probabilistic terms can reveal previously hidden
perspectives. For example, a Gaussian distribution is not appropriate for
modelling discrete observations (implicit clicks or explicit ratings) because its support is
$\mathbb{R}$ and it places zero probability mass at any discrete point.
Furthermore, probabilistic models come with powerful inference
techniques rooted in Bayesian statistics. Machine learning using
probabilistic models (graphical models) can be understood as a
sequence of three steps: 1) posit a generative model of how the
observations are generated; 2) condition on the observed data to infer
posterior distributions of model's unobserved (latent) variables; 3) use
the posteriors according to the generative model to predict unobserved
datum. In the context of matrix factorization, the Bayesian approach
has been shown to produce superior predictions because it can average over many explanations of the data~\citep{SalMnihICML08}.

Poisson matrix factorization (PF) \citep{Gopalan:2013b} addresses the mismatch between
Gaussian models and the discrete observations.
Poisson factorization assumes clicks are Poisson distributed\footnote{In this context maximizing a Poisson likelihood is
equivalent to minimizing the generalized-KL between the user-item matrix
and its approximation.} and user/item factors are gamma distributed. 
This is another difference with respect to Gaussian factorization where factors are typically Gaussian distributed and can therefore 
be negative. Additively combining non-negative factors can lead to
models that are easier to interpret \cite{Seung99}.
Poisson factorization has been shown to outperform several types of matrix
factorization models, including Gaussian matrix factorization, on
recommendation tasks \cite{Gopalan:2013b}. Furthermore, properties of the
Poisson distribution allows the model to easily scale to massive real-life
recommendation datasets.

\mypar{Dynamic Poisson Factorization.}
Poisson matrix factorization models users and items statically over time.
This is unrealistic. 
For example, the preferences of the user in Figure \ref{fig:user755} change from 
from ``Graph Theory'' and ``Quantum Cryptography'' to ``Compressed
Sensing''. Similarly, the readership of the item in Figure \ref{fig:user755}
evolves from ``Graph Theory'' to ``Quantum \& Computational Complexity'' and ``Data
Structures \& Probability''. To be able to capture both the changes of users and
items over time, we introduce dynamic Poisson factorization (dPF).

DPF is a model of user-item clicks over time where time is assumed to be
discrete (i.e., the modeler chooses a time window that
aggregates clicks in chunks of 1 week, 1 month etc.). The aim is to 
capture the evolving preferences of users as well
as the evolution of item popularity over time. Specifically, dPF models a
set of users clicking, or rating, a set of items over time.\footnote{In the rest of
the exposition we will refer to clicks and ratings interchangeably.}

We first give a high-level overview of dPF. The strengths of user preferences and item popularities
are assumed to be a combination of static and dynamic processes. The
static portion assigns each user and item a time-independent
representation, while the dynamic portion gives each user and item a time
evolving model.  The static representation is
the same as in traditional matrix factorization approaches and captures the
baseline interest of users and the baseline popularity of items, which is
then modified by dynamic variables. In this respect, dPF is a
generalization of PF \citep{Gopalan:2013b}.

We now give the mathematical details for how to capture this intuition in a
model. To model time probabilistically, statisticians typically use Gaussian time
series models called state space models \citep{bishop:2006:PRML}. In the
simplest case, state space models draw the state at the next time step from a
Gaussian with mean equal to the previous time step and fixed variance
$\sigma^2$:
\begin{align*}
x_{t} | x_{t-1}  \sim \mathcal{N}(x_{t-1}, \sigma^2).
\end{align*}
We use state space models as the dynamic portion of dPF, where both
the users and items evolve. Formally,
$u_{nk,t}$, the $k$th component of the $n$th user at time $t$ is 
constructed as
\begin{align*}
u_{nk,t} | u_{nk,t-1} \sim \mathcal{N}(u_{nk,t-1}, \sigma_u^2).
\end{align*}
The state space process for items is symmetric. The static components associated
with each user $u_{nk}$ and item $v_{mk}$ are also drawn from a normal distribution. 
They form the intercepts for a time evolving factorization with Poisson observations.
That is, a user's expression of factor $k$ at time $t$ is the sum $\bar{u}_{nk} + u_{nk,t}$. In this sense,
the state space model can be viewed as governing \emph{correction factors} and thus capture the evolution of users'
preferences through time, while static global factors capture the interest
of users that are not influenced by time. 

One issue in using Gaussian state space models is that Poisson observations 
have non-negative parameters (rates). Thus, we exponentiate 
the state-space models intercept sums for each user and item
before combining them to form the mean of the observed rating.
Concretely, the rating for user $n$, item $m$ at time $t$, $y_{nm,t}$
has the following distribution
\begin{align*}
y_{nm,t}\sim\text{Poisson}(\sum_{k=1}^K e^{(u_{nk,t} + \bar{u}_{nk})} e^{(v_{mk,t} + \bar{v}_{mk})}).
\end{align*}
Putting this
all together, the observations are
drawn as follows:  
\begin{enumerate}
\item Draw user global factors: $\bar{u}_{nk} \sim \mathcal{N}(\mu_{\bar{u}},\sigma_{\bar{u}}^2)$
\item Draw item global factors: $\bar{v}_{mk} \sim \mathcal{N}(\mu_{\bar{v}},\sigma_{\bar{v}}^2)$
\item For each time step: $t=1\ldots T$
\begin{itemize}
\item[] \hspace{-0.25in}Draw user and item correction factors: \item[] \hspace{-0.20in}if $t=1$
\item[] \hspace{-0.20in}\hspace{0.1in} $u_{nk,1} \sim \mathcal{N}(\mu_u,\sigma_u^2)$
\item[] \hspace{-0.20in}\hspace{0.1in}  $v_{mk,1} \sim \mathcal{N}(\mu_v,\sigma_v^2)$  
\item[] \hspace{-0.20in}else
\item[] \hspace{-0.20in}\hspace{0.1in}   $u_{nk,t}|u_{nk,t-1} \sim \mathcal{N}(u_{nk,t-1}, \sigma_u^2 )$ 
\item[] \hspace{-0.20in}\hspace{0.1in}  $v_{mk,t}|v_{mk,t-1} \sim \mathcal{N}(v_{mk,t-1}, \sigma_v^2 )$  
\item[] \hspace{-0.25in}Draw a click: 
\item[]   \hspace{-0.20in}$y_{nm,t}\sim\text{Poisson}(\sum_{k=1}^K e^{(u_{nk,t} + \bar{u}_{nk})} e^{(v_{mk,t} + \bar{v}_{mk})})$
\end{itemize}
\end{enumerate}

Unlike Poisson matrix factorization \cite{Gopalan:2013b} our model uses
Gaussian-distributed (global) user and item factors. We could have also used
Gamma-distributed factors \cite{acharya15} although the parametrization of
Gaussians allows us to independantly control the mean and variance of each
Gaussian in the chain.  Further, even with this change, dPF can still model the
long-tail of users and items, an important advantage of Poisson factorization over
Gaussian matrix factorization \cite{Gopalan:2013b}.

\mypar{Analysis and Predictions.} \noindent
We analyze data through the posterior, $p(u, \bar{u}, v, \bar{v} | Y)$. We use
$u$ to represent the set of all user factors and $Y$ the observations. The posterior distribution places probability mass on the latent variables in
proportion to how well they explain the observations. The posterior of
standard Poisson factorization finds a single set of preferences and
attributes, while the dPF posterior places high probability on the sequences of
preferences and attributes that best describe the observed data. Figure~\ref{fig:user755}
plots the expected value under the posterior distribution of the expression of factor
$k$ by user $n$, $u_{nk, t} + \bar{u}_{nk}$, 
and the similar posterior expectation for the items.

In recommendation applications, one wants to use the model to predict the value
of unobserved clicks. We obtain predictions by taking the inner product of the expected
value of the user factors and item factors under the posterior
distribution:
\begin{eqnarray}
\text{E}[y_{nmt}|Y] \approx \sum_k \text{E}_\textrm{posterior} [e^{(u_{nk,t} + \bar{u}_{n,k})} e^{(v_{mk,t} + \bar{v}_{m,k})}]. \label{eq:predictive}
\end{eqnarray}
In practice, recommender systems require rankings (e.g., to show users their top 10 most recommended items when they log on to a website, or to auto-play the most recommended video when the current video has finished). To calculate rankings we use the predictive distribution in Equation~\ref{eq:predictive} as a score for each unobserved rating triplet (i.e., higher predictive probability results in a higher score) and sort the triplets by these scores.

\mypar{Computation.} \noindent
We approximate the posterior distribution with variational inference, which
transforms the posterior estimation problem into an optimization problem. In 
Appendix \ref{sec:inference} we provide the variational inference algorithm for dPF.
The per iteration complexity of
performing posterior inference for dPF is $O(T(R + NK + MK))$ where $R$ is the
number of non-zero observations of the most populous time step ($R=\max_t
R_t$), $T$ the number of time steps, $N$ and $M$ the number of users and items,
and K the number of latent components respectively. 

The updates of each factor depend only on the other factors in its \emph{Markov
blanket} (a factor's parents, its children, and its children's parents).
As a consequence, the updates to the variational parameters of
each user are independent of all other users (and similarly for items).
Hence we can easily parallelize the posterior inference computations.
We implement dPF in C++ and use openMP for parallelization.\footnote{Code
is available: \url{https://github.com/Blei-Lab/}} Empirically, we obtain a
near linear speedup in the number of threads. 

\mypar{Implementation Details.} \noindent We initialize hyperparameters 
($\mu_u,\mu_v,$ $\bar{\mu}_u,\bar{\mu}_v$) controlling the means to the priors to small random numbers close to 0 and
variance hyperparameters
($\sigma_u,\sigma_v,\sigma_{\bar{u}},\sigma_{\bar{v}}$) to 10. We found that
these values work well across different datasets. We provide indications that
the model is well-behaved under a wide range of hyperparameter settings
(Section~\ref{sec:hyperparams}). We have found that for interpretability, it is
important to use the same initialization for factors across time. We
experimented with initializing the global factors with PF but that did not
provide an advantage. We ran the the numerical optimization routines for up to
500 iterations or until convergence. 
 
\makeatletter{}\section{Experiments}\label{sec:experiments}

\noindent
We now fit our model using datasets consisting of users consuming movies and
scientists clicking on papers. As we describe in Section \ref{sec:model} we can
use the expectation of the posterior distributions over user and item latent
factors to provide item recommendations to users through time. In addition to
measuring the performance of dPF against baselines we gain insights into the
datasets and the model by exploring its posteriors. We highlight the following
results:
\begin{enumerate}
\item dPF provides better recommendations across multiple datasets compared to
several static and dynamic baselines.
\item dPF scales to large datasets.
\item dPF allows us to explore and gain interesting insights into the arXiv
dataset. 
\end{enumerate}

In Section~\ref{sec:datasets}, we describe the two datasets used in our
experiments. We then formalize the experimental methodology and the comparative
metrics in Section~\ref{sec:baselines}. Finally, we provide comparative and
explorative results in Sections~\ref{sec:results_compare},~\ref{sec:more_results}, and~\ref{sec:exploration}.

\subsection{Datasets}
\label{sec:datasets}
\noindent
We study two different datasets. 

\emph{Netflix-time:} First, we consider a movie
dataset derived from the Netflix challenge. We follow a similar procedure as
described in~\citet{Li:2011:CCF:2283696.2283780} to obtain a subset of the
original Netflix dataset focusing on users and movies that have been active
over several years. The resulting dataset contains 7.5K users, 3.6K items, and
2 million non-zero observations over 14 time steps (with a granularity of 3
months). The validation and test sets are composed of all the clicks occurring
in the final 9 months of the dataset (90K clicks). Since our focus is on the
implicit data case, we binarize the data such that every observed rating is a
``1'' rating and all unobserved ratings are zeros. 

\emph{Netflix-full:} In addition to the Netflix-time dataset, we fit dPF on the whole
Netflix dataset (with a granularity of 6 months) where we kept the last 1.5
years of click for validation and test. The resulting dataset contains over
225K users and 14K items and 6.9M observations. This dataset is more
challenging for inference because it includes items in the long tail (i.e., a
large number of items used by very few users) and very inactive users.

\emph{arXiv:} DPF is also fit on a click data from the arXiv.org site.
arXiv.org is a pre-print server for scientific papers. We have access to ten
years of clicks from registered users of arXiv. We keep 75K papers with at
least 20 clicks over the ten years and 5K users with clicks spanning at least
five years (1.3M observations). We set the time granularity to be six months
which seems reasonable to capture scientists' changing preferences. We verified
that the results are similar with other time periods. The resulting dataset
contains 19 time periods. We keep all observations of the last time step (the
last six months) for validation and testing. 

\emph{arXiv-cs-stats:} Physicists were early adopters of the arXiv. As such,
fits to the above dataset are dominated by that field. To gain insights into
our model's discoveries we have found it useful to restrict our attention to
subject areas that were closer to our computer science expertise. This dataset
is a cut of the data with papers that are in either computer science (CS) or
statistics (Stats) fields. The field(s) of each paper is selected by its
authors when submitting. The resulting data has just under 1K users, 5K
items and spans 7 years worth of clicks (58K observations).  We use this data
for explorative purposes (Section~\ref{sec:exploration}). 

\subsection{Baselines}
\label{sec:baselines}
\noindent
We are able to compare to \emph{BPTF}, a recently-proposed dynamic model based
on tensor factorization approach. \citet{lxiong:10:bptf} model the changing
strengths of factors over time with a probabilistic tensor approach. In
more detail, the observation matrix is assumed to decompose into a tensor
product between user preferences, item popularities, and temporal trends. Their
approach captures aggregate changes in behavior but is not specific enough to
model the time-dependent shifts of individual users or items. They use Gibbs
sampling to infer the latent factors given explicit data only, an approach that
does not scale when considering implicit data on larger datasets. To adapt this
method to the implicit data case we sampled a random subset of unobserved
items such as to observe an equal number of zeros and ones.  

Another natural benchmark to demonstrate the benefits of time modelling is Poisson
matrix factorization, since our model builds on that approach. 

An effective time model should be able to learn how to weight the different
clicks based on recency and retain the information from older clicks only to
the extent that it is useful to predict new ones.  For example, in a system
where user preferences quickly evolve, the system may rapidly forget about
older preferences. To clearly demonstrate the benefits of time modelling we
compare to two different PF models. The first is trained using all past ratings
(\emph{PF-all}) while the second is trained using only observations at the last
(train) time period (\emph{PF-last}). 

We note that the implicit data case is significantly more challenging than
the usual explicit data case. First, the observations are extremely
imbalanced. For example in the datasets used in the experiments ``0''
observations outnumber ``1'' observations by a ratio of over 250 to 1 (arXiv). 
Second, implicit data methods train on exactly $NM$ observations. Finding
baselines that will scale to these data sizes is challenging. Furthermore, our
focus on implicit data means that we cannot readily compare to previously
published results.  

We also fit the popular TimeSVD++ model
\citep{DBLP:journals/cacm/Koren10}. It is not designed for the implicit data
case and preliminary experiments (where we studied different methods
of incorporating zero observations) on our Netflix-time dataset showed that it
performed significantly worse than all baselines (including both static PF
models).

For all experiments we set the number of components to $K=20$. Our results
generalize beyond this value.

\subsection{Click prediction \& Metrics}
\label{sec:results_compare}
\noindent
We simulate a realistic scenario where each method is given access to past user
clicks and the models must predict items of future interest to the users.
Further for each dataset we repeat this process for each time step. That is for
each time step $t$ we train with all observations before $t$ (i.e., $1\ldots
t-1$) and evaluate predictions on the observations at $t$. When reporting results
we average over all time steps.

Evaluating the accuracy of a collaborative filtering model for implicit data is
usually done using ranking-related error metrics.  We consider user recall
given K top-items to recommend (``recall@K'' see below for details). 
However, we notice that on datasets
containing many items, models often fail to place \emph{any} of the user's
consumed items in the top K. These metrics limit our ability to understand
the merits of different models.
As a remedy we turn to metrics which take into account all
test items while discounting the contribution of an item as a function of
its predicted rank.  The popular NDCG, and mean reciprocal rank (MRR) both
have this property.  MRR linearly decays the relevance of an item as a
function of its rank while NDCG logarithmically decays them. We also report
the mean average rank (MAR) metric.

Formally, for binary data the four chosen metrics are: 
\begin{itemize}
\item[] \hspace{-0.15in} - Recall@T: $\frac{1}{N} \sum_{i=1}^N \sum_{j\in
\in \mathbf{y}_i^U} \frac{\mathbf{I}(\text{rank}(i,j)\leqslant T)}{\text{min}(T,|\mathbf{y}_i^U|)}$ \\
where $\mathbf{I}$ is the indicator function. In our experiments we set T to 50
for all datasets which is a reasonable number of items to present to a user.
Results are consistent across other values of T that we have studied. 
\item[] \hspace{-0.15in} -  Normalized discounted cumulative gain (NDCG): \\$\frac{1}{N} \sum_{i=1}^{N} \sum_{j\in \mathbf{y}_i^U} \frac{1}{\log(\text{rank}(i,j) + 1)}$.
\item[] \hspace{-0.15in} -  Mean reciprocal rank (MRR): $\frac{1}{N} \sum_{i=1}^{N} \sum_{j\in \mathbf{y}_i^U} \frac{1}{\text{rank}(i,j)}$.
\item[] \hspace{-0.15in} -  Mean average rank (MAR): $\frac{1}{N} \sum_{i=1}^{N} \sum_{j\in \mathbf{y}_i^U} \text{rank}(i,j)$.
\end{itemize}
$\mathbf{y}_i^{U}$ is the set containing user $i$'s test observations and $N$ the number of
users. $\text{rank}(i,j)$ is the predicted rank of item $j$ for user $i$. 

\subsection{Results} \label{sec:more_results}
\noindent The results across all datasets consistently show the superior
performance of dPF according to all baselines. Below we examine these results
in more detail. 

 We first turn our attention to the Netflix data. On the Netflix-time
dataset dPF outperforms all baselines according to all metrics (Table
\ref{tab:netflix_2}). We note that the BPTF baseline mostly outperforms PF-all
and PF-last. This indicates that while BPTF's time model is useful, modeling
users' and items' individual evolutions is preferable. 

The performance on Netflix-full is similar where dPF also outperforms all
baselines. Other authors have shown that outperforming static models on Netflix
is challenging \citep{lxiong:10:bptf}.  As outlined in
\citet{DBLP:journals/cacm/Koren10}, Netflix users who change at a very fine level of
granularity (daily) may be hard to capture with a coarser granularity.
Furthermore, we note that users who rarely use the service do not provide
enough data to disambiguate between a small number of changing user interests
and a wider number of \emph{static} user interests that happen to be exhibited
over many time steps (a similar argument applies to movies in the long tail).
Nonetheless dPF is able to slightly outperform static methods by capturing
evolving preferences and viewer ship where available.

Fitting BPTF to this larger dataset is prohibitively expensive even with our
proposed subsampling approach. As such we could not report results on the
larger Netflix dataset nor on the arXiv dataset.

The arXiv is a good testbed for our model since scientists and science evolve
and we believe that it may be possible to capture such evolutions using arXiv's
click data. We note that dPF outperforms all other methods on arXiv (Table
\ref{tab:arxiv_2}). This further validates that dPF provides better
recommendations by modelling the evolution of users and items. 

From a methodological point of view it is also interesting to note that there
is not always perfect correlation between all the metrics (that is the rank of
the baselines change under different metrics). Amongst them MAR seems to be
noisiest metric. It can probably be explained by the fact that it is less
robust to outliers because it weights every test observation evenly.

\begin{table}[h]
\begin{center}
\begin{tabular}{p{0.75in}cccc}
\hline
           & Recall@50 & MAR & MRR & NDCG \\ \hline 
dPF        & {\bf 0.170} & {\bf 640} & {\bf 0.027} & {\bf 0.294} \\ 
BPTF \citep{lxiong:10:bptf}   & 0.148  & 668 & 0.020  & 0.277 \\ 
PF-all \citep{Gopalan:2013b}  &  0.145 & 691 & 0.021 & 0.280 \\ 
PF-last \citep{Gopalan:2013b} & 0.065 & 807  & 0.019 & 0.268 \\ \hline
\end{tabular}
\caption{Performance of dPF versus baselines on Netflix-time. Bold numbers
indicate the top performance for each metric. We were unable to obtain
competitive performance from TimeSVD++\cite{DBLP:journals/cacm/Koren10}.}
\label{tab:netflix_2}
\end{center}
\end{table}

\begin{table}[h]
\begin{center}
\begin{tabular}{p{0.75in}cccc}
\hline
             & Recall@50 & MAR & MRR & NDCG \\ \hline 
dPF          &  {\bf 0.156} & {\bf 1605} & {\bf 0.021 } & {\bf 0.358} \\ 
PF-all~\citep{Gopalan:2013b} & 0.120  & 1807 & 0.015 & 0.338 \\ 
PF-last~\citep{Gopalan:2013b} & 0.138 & 1635 & 0.018 & 0.351 \\ \hline
\end{tabular}
\caption{Performance of dPF versus PF on the Netflix-full dataset. 
Bold numbers indicate the top performance for each metric. While dPF does best
according to all metrics, PF-last slightly outperforms PF-all. The performance
of PF-last seems to indicate that user preferences and movie popularities vary
rapidly.}
\label{tab:netflix_full}
\end{center}
\end{table}

\begin{table}[h]
\begin{center}
\begin{tabular}{p{0.6in}cccc}
\hline
           & Recall@50 & MAR   & MRR    & NDCG \\ \hline 
dPF        & {\bf 0.035} & {\bf 21822}  & {\bf 0.0062} & {\bf 0.186} \\ 
PF-all  \citep{Gopalan:2013b} & 0.032 & 22402 & 0.0056 & 0.182 \\ 
PF-last \citep{Gopalan:2013b} & 0.023 & 25616 & 0.0040 & 0.168 \\ \hline
\end{tabular}
\caption{Performance of dPF versus baselines on arXiv. Bold numbers indicate
the top performance for each metric.}
\label{tab:arxiv_2}
\end{center}
\end{table}

\mypar{Effect of hyperparameter values.}
\label{sec:hyperparams}

\begin{figure}
\begin{center}
\includegraphics[width=0.75\linewidth]{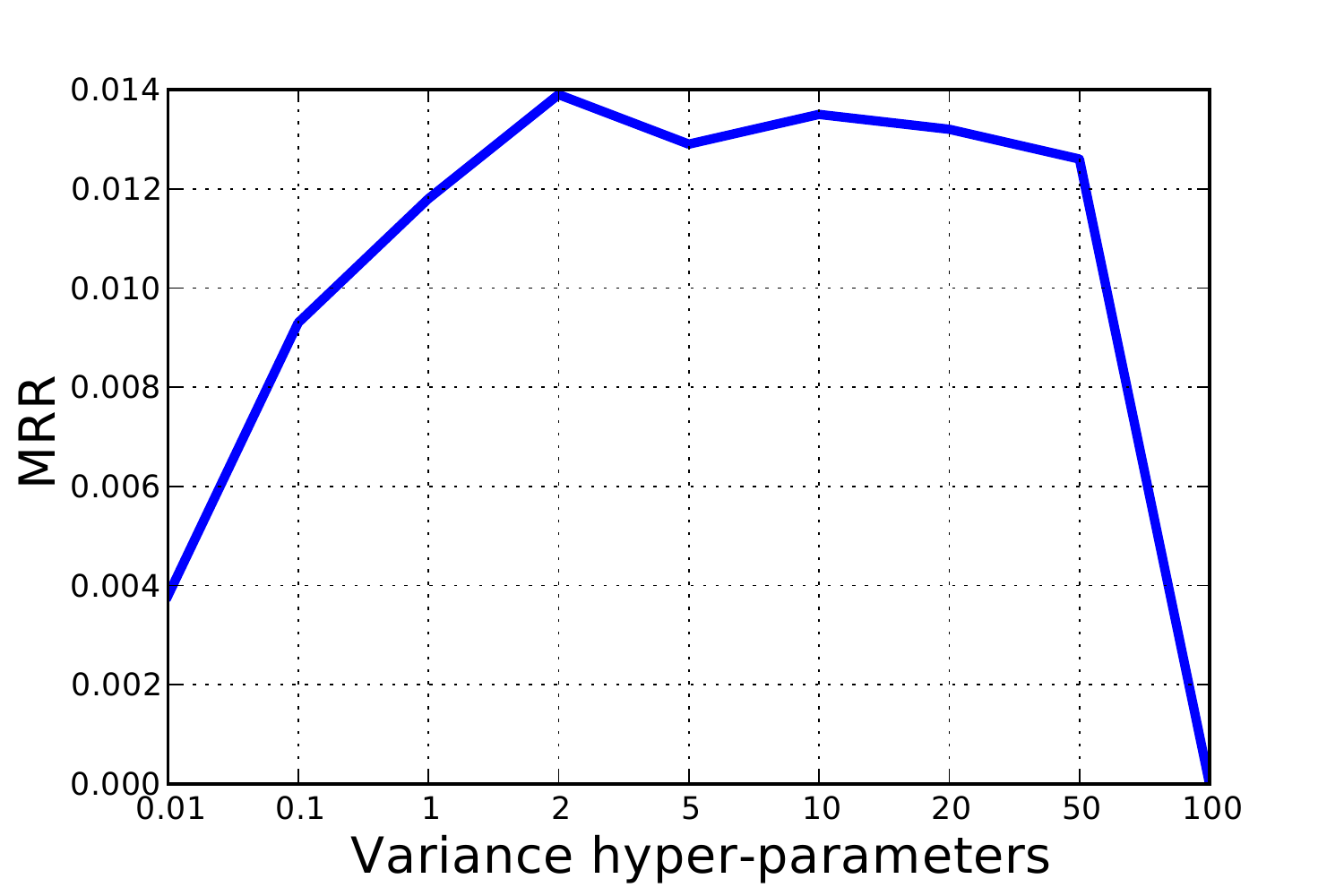}
\end{center}
\caption{dPF performs well across a large range of variances (2--50) by MRR in a pilot study of a single timestep of the arXiv data.}
\label{fig:mrr_vs_variance}
\end{figure}
\noindent
We have found that dPF is fairly robust to the values of its hyperparameters.
We demonstrate it more formally for the variance hyperparameters (Figure
\ref{fig:mrr_vs_variance}). We note that except at the extremes the performance
of dPF is good for a wide-range of variances. In our experimental setup the
last time step is used for predictions. This can may explain the model's
robustness.

\begin{figure}
\begin{center}
\includegraphics[width=0.5\textwidth]{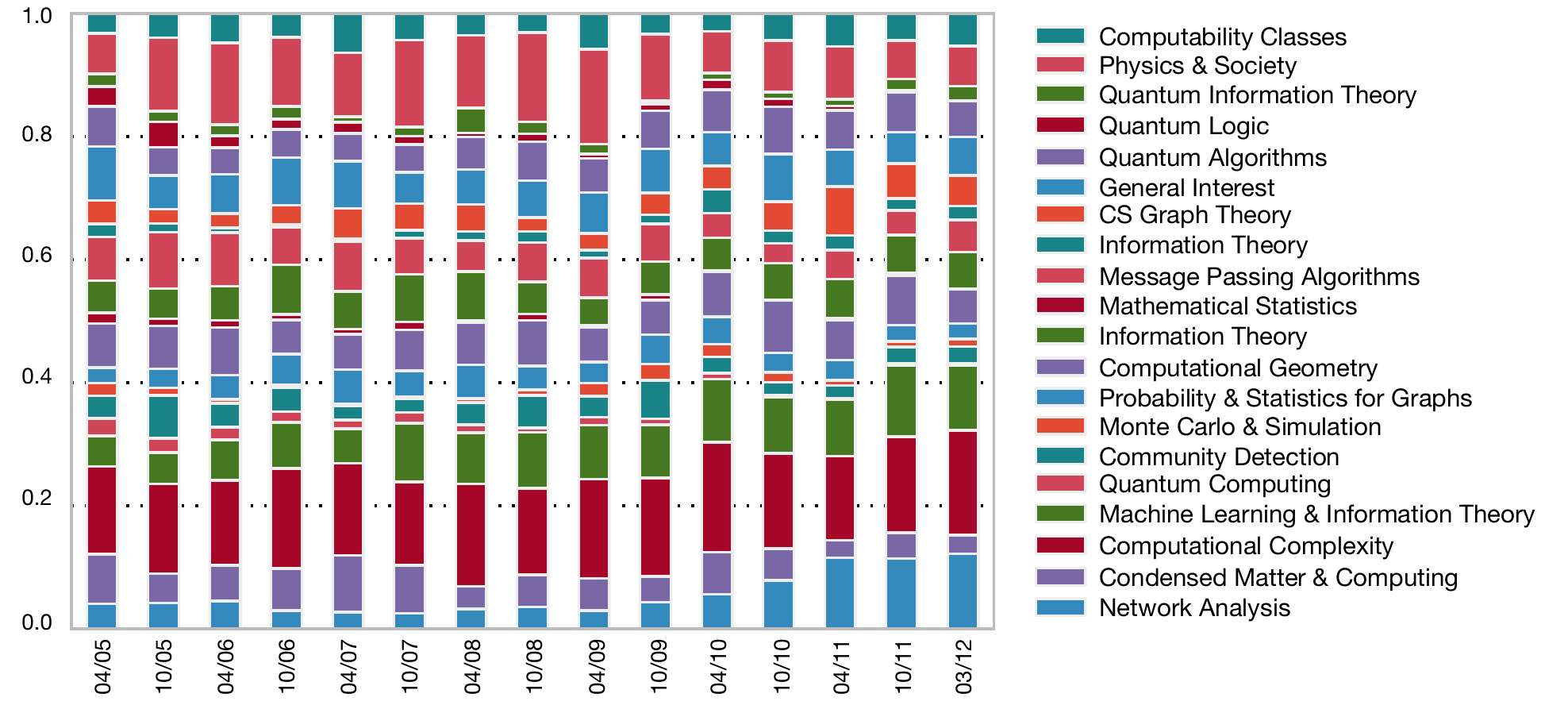}
\caption{Evolution of latent factors, corresponding to research areas, as discovered by dPF from the raw click data.
We see that network analysis has been a growing area of research in computer science over the past decade.}
\label{fig:topics_over_time}
\end{center}
\end{figure}

\begin{figure}
\begin{center}
  \includegraphics[width=\linewidth]{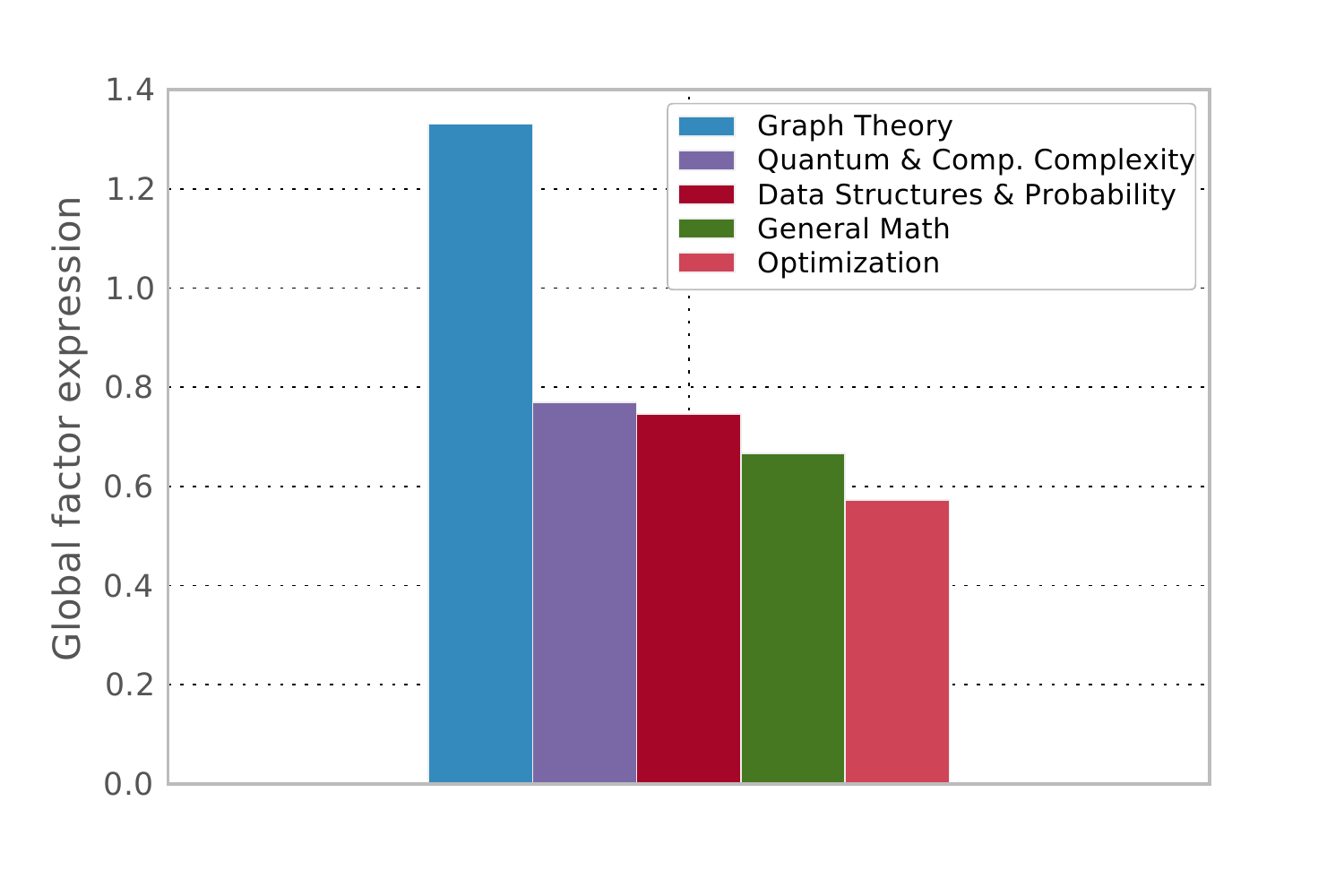}
\end{center}
\caption{The expression level of five static factors of ``The Google Similarity Distance'' paper. When compared to the dynamic trend in Figure~\ref{fig:user755},
we see that the static factors identify the paper's main field, graph theory. 
Moreover, dPF can use the local factors to express excursions to other fields.
}
\label{fig:item4663_global}
\end{figure}

\subsection{Exploration} 
\label{sec:exploration}
\noindent
There are various ways of evaluating a probabilistic model. While we have
provided conclusive empirical results above we now turn to exploration of the
model's posteriors. Exploring the posterior can be a useful diagnosis tool
\citep{doi:10.1146/annurev-statistics-022513-115657}. Here our aim is to
further validate our modelling choices. We also want to better understand
the arXiv-cs-stats dataset. 

In Figure \ref{fig:user755} we showed, using the local expected value of its
factors under their posterior distribution, an item that garnered interest
from different communities over time. We can also examine the model's global
expected value under the posterior distribution (Figure
\ref{fig:item4663_global}). In addition to the four factors in Figure
\ref{fig:user755} we added an extra factor (labeled ``Optimization'') which
is unused by this paper. The ``Graph Theory'' factor is the highest
global factor, which is sensible since it expresses the paper's main field. The
other three factors are expressed at levels which are in between ``Graph
Theory'' and ``Optimization''. The global popularity of this paper in these
four fields is the same. In particular, dPF can simply use local factors to
explain ephemeral popularity in other fields.

In addition to looking at single users and items we can explore the model at the
aggregate level to discover global patterns. For example, we can look at the
evolution of different scientific fields over time. We plot the expected value
of all 20 factors inferred using the arXiv-cs-stats dataset (Figure
\ref{fig:topics_over_time}). We corrected for
arXiv's growing popularity over time.  Scientific fields on arXiv are all well
established, and as such, their popularities are stable over time. Note that this does
not mean that fields do not evolve. We also notice that according to dPF, some
fields have gained in popularity (in particular ``Machine Learning \&
Information Theory'' as well as ``Networks \& Society'').

\makeatletter{}\section{Conclusions \& Future Work} 

\noindent
We examined the problem of modelling user-item clicks through time.  Our main
contribution is the dynamic Poisson factorization model (dPF). In addition
to modeling both users and items changing through time and being able to scale to large datasets 
using implicit information, we
showed the empirical advantages of modelling the evolution of users and items.
Importantly we used this model to explore and better understand the arXiv, an 
indispensable tool to many scientific fields.

There are several opportunities to extend this work. It would be
interesting to model aggregate user trajectories to better understand how
(scientific) communities evolve. It would also be worthwhile to allow users and
items to be modelled at multiple different granularities. This would allow a
model to understand both short and long-term evolutions. Although we can set
the granularity in our current computation (and memory consumption) may become
an issue as the granularity gets small.  Continuous time models may offer
interesting avenues to avoid specifying granularities.

\mypar{Acknowledgements.} We thank the reviewers for their helpful
comments. RR is supported by an NDSEG fellowship. DMB is supported by NSF
IIS-0745520, NSF IIS-1247664, ONR N00014-11-1-0651, and DARPA FA8750-14-2-0009.

{
\bibliographystyle{abbrvnat}
\bibliography{dpf}  
}

\newpage
\appendix
\makeatletter{}\section{Inference}\label{sec:inference}

\noindent
Performing Bayesian inference on the random variables of dPF requires calculating the posterior distribution. This is difficult due to the non-conjugacy (lack of analytic integrals) between the log-normal
distribution of the factors and the Poisson-distributed observations.
We therefore resort to approximate inference techniques (e.g., Markov chain Monte Carlo or variational methods).

Variational inference frames the posterior inference task as the
problem of minimizing the KL-divergence
between an approximate distribution and the true posterior. This
optimization approach is orders of magnitudes faster than sampling-based
approaches, at the cost of not being able to capture the full Bayesian
posterior. 
We assume that our approximating family (in variational inference) fully
factorizes. The variational distribution of individual factors is
normally-distributed:
\begin{eqnarray}
  q(\mathbf{u}_{nt}) &=& \prod_k q(u_{nt,k}|\hat{\mu}_{unt,k},\hat{\sigma}_{unt,k}^2) \nonumber \\
  &=& \prod_k
  \mathcal{N}(u_{nt,k}|\hat{\mu}_{unt,k},\hat{\sigma}_{unt,k}^2) \nonumber
\end{eqnarray} 
We treat the other time-independent ($v$) and time-dependent ($\bar{u}$ and
$\bar{v}$) latent variables in a similar way. To differentiate the variables of the generative model from
their variational counterparts we denote the latter with a circumflex (a
``hat'').

It is straightforward to show that minimizing the KL-divergence between
the q distribution and the true posterior is equivalent to maximizing a lower bound on the
model's log evidence~\citep{Jordan:1999:LGM:308574}.\footnote{The model evidence is also referred to as the
marginal likelihood.} We use the bold symbols to denote sets of factors, for example the
set of all user factors: $\mathbf{u} \equiv
\{ u_{nt,k} \}_{n=0:N,t=1:T}$. $\zeta$ stands for the set of all
hyperparameters: $\{\mu_u,\sigma_u,\mu_v,\sigma_v,\mu_{\bar{u}},\sigma_{\bar{u}},\mu_{\bar{v}},\sigma_{\bar{v}}\}$. $\lambda$
denotes all of the parameters to the variational distribution such as $\hat{\mu}_{unk,t}$ and $\hat{\sigma}_{unk,t}^2$.
For our model, this evidence lower bound~(ELBO) is
\begin{align}
 p(\mathbf{y} | \zeta ) & \geqslant   {\cal L}(\lambda) \\ 
                        & := \text{E}_q[ \log
  p(\mathbf{y}, \mathbf{u},\mathbf{v},\mathbf{\bar{u}},\mathbf{\bar{v}} | \zeta )   ] 
        - \text{E}_q [ \log q(\mathbf{u},\mathbf{v},\mathbf{\bar{u}},\mathbf{\bar{v}}) ].\nonumber 
\label{eq:elbo}
\end{align}

The fully-factorized approximation allows us to expand the ELBO as follows:
{\small
\begin{align*}
& \sum_{n}^N \E_{q}[\log p(\mathbf{\bar{u}}_{n} | \bm{\mu}_{\bar{u}}, \sigma^2_{\bar{u}}\mathbf{I}) ] 
+ \sum_{m}^M \E_{q}[\log p(\mathbf{\bar{v}}_{m} | \bm{\mu}_{\bar{v}}, \sigma^2_{\bar{v}}\mathbf{I}) ] \\ 
&+ \sum_{t=2,n}^{T,N} {\E_{q}[ \log p(\mathbf{u}_{n,t} | \mathbf{u}_{n,t-1}, \sigma_u^2\mathbf{I}) ]} + \sum_{n}^N{\E_q[\log p(\mathbf{u}_{n,1} | \bm\mu_u, \sigma_u^2\mathbf{I}) ]} \\ 
&+ \sum_{t=2,m}^{T,M} {\E_{q}[ \log p(\mathbf{v}_{m,t} | \mathbf{v}_{m,t-1}, \sigma_v^2\mathbf{I})]} + \sum_{m}^M{\E_q[\log p(\mathbf{v}_{v,1} | \bm\mu_v, \sigma_v^2\mathbf{I}) ]} \\
&+ \sum_{n,m,t}^{N,M,T} {\E_{q}[\log p(y_{nm,t} | \sum_k exp(u_{nk,t} + \bar{u}_{nk})^T exp(v_{mk,t} + \bar{v}_{mk})) ]} \\
& + {H(q)} 
\end{align*}}
where $H(q)$ is the entropy of the variational distribution.

To optimize the ELBO one typically expands the expectations in
Equation~\ref{eq:elbo}. The variational parameters can then be optimized using coordinate or gradient ascent. The non-conjugacy in our model implies that further
manipulations are required to express the expectations from
Equation~\ref{eq:elbo} in closed form. In particular 
the term requiring an additional approximation comes from the
Poisson distribution. The Poisson pdf, with
rate $\lambda$ is $\frac{\lambda^k}{k!} e^{-\lambda}$. 
The first term in the ELBO as expressed in Equation \ref{eq:elbo} (the joint distribution over latent and
observed variables) can be re-written using the product rule of probability as:
\begin{align*}
\text{E}_q[ \log p(\mathbf{y}|\mathbf{u},\mathbf{v},\mathbf{\bar{u}},\mathbf{\bar{v}}) + \log p(\mathbf{u},\mathbf{v},\mathbf{\bar{u}},\mathbf{\bar{v}} | \zeta ) ].
\end{align*}
The first term corresponds to the Poisson distribution. It can be expanded into: 
\begin{align*}
\text{E}_q[ \log \frac{\lambda^k}{k!}\exp^{-\lambda} ] 
&= \text{E}_q[ k\log\lambda - k! -\lambda ] \\ 
&= \text{E}_q[ k\log\lambda ] - \text{E}_q[ \lambda ] - k!
\end{align*}
where $\lambda = \sum_k \exp(u_{nk,t} + \bar{u}_{nk}) \exp(v_{mk,t} +
\bar{v}_{mk} )$. We do not know of a closed-form exact solution to the first
expectation when $k$ is non-zero.
Fortunately several known techniques can be used to approximate this
integral~\citep{Wang:2013}. We re-use an idea from the recent
Poisson factorization which consists of adding a set of auxiliary random
variables. For dPF, this means adding auxilliary variables
$\phi_{mnt,k}$ one for each user-item-timestep component for all of 
the non-zero ratings. Formally,
this allows for the following manipulation: 
\begin{align}
&\log \sum_k \exp(u_{nk,t} + \bar{u}_{nk}) \exp(v_{mk,t} + \bar{v}_{mk} )  \\ 
&= \log\sum_k \frac{\phi_{nmk,t}}{\phi_{nmk,t}} \exp(u_{nk,t} + \bar{u}_{nk}) \exp(v_{mk,t} + \bar{v}_{mk} ) \\ 
&= \log \text{E}_{\mathbf{\phi}} [ \frac{\exp(u_{nk,t} + \bar{u}_{nk})
\exp(v_{mk,t} + \bar{v}_{mk} )}{\phi_{nmk,t}} ] \\ 
&\geqslant \text{E}_{\mathbf{\phi}} [ u_{nk,t} + \bar{u}_{nk} +
v_{mk,t} + \bar{v}_{mk} - \log\phi_{nmk,t} ] \\ 
\label{eq:double-jensen}
\end{align}
We obtain the last inequality by applying Jensen's inequality. We note that,
unlike in Poisson factorization, the resulting model is not conditionally
conjugate, which intuitively means the distribution of a variable given everything else 
has known form.
As a result, the equations do not yield closed-form updates.
We use numerical optimization and perform coordinate ascent on the ELBO
using a quasi-Newton optimizer (L-BFGS). The update for $\phi_{nmt,k}$ can
be solved for in expected value with respect to $q$ of Eq. \ref{eq:double-jensen} as
\begin{align}
\phi_{nmk,t} \propto \exp(\hat{u}_{nk,t} + \hat{\bar{u}}_{nk} +
\hat{v}_{mk,t} + \hat{\bar{v}}_{mk})
\label{eq:phi_up}
\end{align}

The algorithm used to infer the
variational posteriors (i.e., train the model) is shown in
Figure~\ref{fig:algo}.

\begin{figure}
\begin{itemize}
\item[] For $t=1\ldots T:$
\begin{itemize}
\setlength{\itemindent}{-.2in}
  \item[] Update $\hat{\mu}_{unk,t} \text{ and } \hat{\sigma}_{unk,t} \text{with L-BFGS using} \nabla_{\hat{\mu}_{unk,t}, \hat{\sigma}_{unk,t}} {\cal L} \;\;\forall n,k$
  \item[] Update $\hat{\mu}_{vmk,t} \text{ and } \hat{\sigma}_{vmk,t} \text{with L-BFGS using} \nabla_{\hat{\mu}_{vmk,t}, \hat{\sigma}_{vmk,t}} {\cal L}  \;\;\forall m,k$
  \item[] Update $\hat{\phi}_{nmk,t} \;\;\forall n,m,k,t \mid y_{n,mt} > 0$ using Eq. \ref{eq:phi_up}
\end{itemize}
\item[] Update $\hat{\mu}_{\bar{u}n,k} \text{ and } \hat{\sigma}_{\bar{u}n,k}  \text{with L-BFGS using} \nabla_{\hat{\mu}_{\bar{u}n,k}, \hat{\sigma}_{\bar{u}n,k}}\;\;\forall n,k$
\item[] Update $\hat{\mu}_{\bar{v}m,k} \text{ and } \hat{\sigma}_{\bar{v}m,k}  \text{with L-BFGS using} \nabla_{\hat{\mu}_{\bar{v}m,k}, \hat{\sigma}_{\bar{v}m,k}} {\cal L} \;\; \forall m,k$
\item[] Repeat until convergence
\end{itemize}
\caption{Coordinate ascent algorithm for dPF. We first update the
(variational) correction factors and subsequently we update the (variational)
global factors. Prior to each update the corresponding vector
$\boldsymbol{\phi}_{nmt}$ must be evaluated.}
\label{fig:algo}
\end{figure}

\end{document}